\RequirePackage{fix-cm}
\documentclass[twocolumn,compsoc]{CVM_}
\graphicspath{{figures/},{figures/cvm/}}
\def\ManuscriptInfo{}

\def\PaperTitle{Photometric stereo for strong specular highlights}

\def\AuthorNames{Maryam Khanian$^1$(\rm{\Letter}), Ali Sharifi Boroujerdi$^1$, and Michael Breu\ss$^1$ \rm }
\def\AuthorAdress{$1\quad $ & Chair of Applied Mathematics, Brandenburg Technical University,
    Cottbus 03046,  Germany. E-mail: Maryam Khanian, khanimar@b-tu.de (\rm{\Letter}); Ali Sharifi Boroujerdi,  shariali@b-tu.de; Michael Breu\ss, michael.breuss@b-tu.de\\
   
}

\def\firstheader{}

\headevenname{{M. Khanian \etal} }

\begin{document}

\MakePageStyle
\frenchspacing
\MakeAbstract{Photometric stereo (PS) is a fundamental technique in computer vision known to produce 3-D shape with high accuracy. 
The setting of PS is defined by using several input images of a static scene taken from one and the same camera position but under varying
illumination. The vast majority of studies in this 3-D reconstruction method assume orthographic projection for the camera model. 
In addition, they mainly consider the Lambertian reflectance model as the way that light scatters at surfaces. So, providing reliable PS results 
from real world objects still remains a challenging task. We address 3-D reconstruction by PS using a more realistic set of assumptions 
combining for the first time the complete Blinn-Phong reflectance model and perspective projection. To this end, we will compare two different 
methods of incorporating the perspective projection into our model. Experiments are performed on both synthetic and real world images. 
Note that our real-world experiments do not benefit from laboratory conditions. 
The results show the high potential of our method even for complex real world applications such as medical 
endoscopy images which may include high amounts of specular highlights.
}

\MakeKeywords{Photometric stereo, high specularity, complete Blinn-Phong model, perspective projection, diffuse and specular reflection, CCD camera}

\section{Introduction}\label{sec:introduction}

The reconstruction of three dimensional (3-D) information at hand of two dimensional images 
is a classic problem in computer vision. Many approaches exist to tackle the task, as documented 
by a rich literature and a number of excellent monographs, among them let us mention here
\cite{BKP1986,TrVe98,Wo2013}. Let us also mention the survey \cite{Ihrke2010} on 3-D reconstruction methods
that may be more oriented towards the computer graphics community.
Following \cite{Wo2013} one may distinguish approaches based 
on the point spread function as in depth from focus or defocus \cite{XiSha1993}, 
triangulation-based methods such as stereo vision \cite{Faugeras1993} or structure from motion
\cite{ToKa1992} and intensity-based or photometric methods such as shape from shading or 
photometric stereo \cite{BKP1986}.
An abundance of specific approaches exist that may be roughly classified at hand of the mentioned proceeding.
Generally speaking these may be distinguished depending on the type of image data, the number of acquired input images, or if 
the camera or objects in the scene may move or not. As an example let us mention here techniques 
based on specular flow \cite{Adatoetal2010,Godardetal2015,Sankaranarayanan2010} relying on relative motion 
between a specular object and its environment. 

Focusing on photometric approaches, as mentioned by Woodham \cite{woodham78} 
and Ihrke \etal {\hspace{0.2ex}} \cite{Ihrke2010}
these typically employ a static view-point and variations in illumination
to obtain the 3-D structure. While shape from shading is the corresponding photometric technique
classically making use of just one input image cf.\ \cite{BKP1986}, 
photometric stereo (PS) allows to reconstruct the depth map of a static scene 
from several input images taken from a fixed view point under different illumination 
conditions. The pioneer of the PS problem was Woodham in 1978 \cite{woodham78},
see also Horn \etal {\hspace{0.2ex}} \cite{HWS78}.
Woodham derived the underlying image irradiance equation as a relation between the image intensity 
and the reflectance map. It has been shown that the Lambertian surface orientation can be uniquely 
determined from the resulting appearance variations provided that the surface is illuminated by at 
least three known, non-coplanar light sources, each for an individual input image \cite{Woodham1980}. 
\par
As it is for instance also recognized in \cite{Ihrke2010},
most of the later approaches have followed Woodham's idea and kept two simplifying assumptions. Of particular importance, the first one 
supposes that the surface reflects the light according to Lambert's law \cite{lambert1760}. This simple reflectance model can still be a 
reasonable assumption on certain types of materials, when the scene is composed of matte surfaces, but fails for shiny objects concentrating 
light distributions. Such surfaces can readily be seen in real world situations. It is quite well proved that a light source illuminating a 
rough surface, reflects a significant part of the light as described by a non-Lambertian reflectance model \cite{Tagare1988,Branden1966,Beckmann1963}. 
In such models the intensity of reflected light depends not only on the light direction but also on the viewing angle, and the light is reflected 
in a mirror-like way accompanied by a specular lobe. The second assumption in classic PS models is that scene points are projected orthographically during the 
photographic process. This is a reasonable assumption if objects are far away from the camera, but not if they are close in which the 
perspective effects grow to be important. The importance of using the perspective projection in such a situation has been demonstrated
in the computer vision literature, in the context of photometric methods let us refer for instance to the work \cite{TSY2005} 
where a corresponding example is discussed in detail.
\par
Many studies in PS considered the non-Lambertian effects as outliers and tried to remove them.
Mukaigawa \etal {\hspace{0.2ex}} \cite{MIS07} suggested a random sample consensus based approach where only diffuse reflection 
is selected from among the candidates.
Mallick \etal{\hspace{0.2ex}} {\cite{Mallick2005}} introduced a rotation transformation for transforming the RGB color channel to a 
SUV color channel with the specular channel S and diffuse channels UV. Then, the specular channel S is used for removing specularities.
Chanki \etal{\hspace{0.2ex}} \cite{CYS10} introduced a strategy based on a maximum feasible subsystem approach.
In their method, the maximum subset of images satisfying the Lambertian constraint is obtained
among the whole set of PS images that include non-Lambertian effects like specularities. A median filtering technique is illustrated by 
Miyazaki \etal {\hspace{0.2ex}} \cite{MHI10} to evade the influence of specular reflections which they considered as outliers.
Another method relying on this concept is presented by Tang \etal {\hspace{0.2ex}} \cite{TTW05} who proposed a coupled 
Markov Random Field based on treating the specularities and shadows as noise. Wu \etal{\hspace{0.2ex}} \cite {WGS10} 
considered the 3-D recovery problem using a convex optimization technique 
for separating specularities as deviations from the 
basic Lambertian assumption in the objective function. Smith and Fang \cite{Smith2016} used a model-based approach that excludes observations that 
do not fit the Lambertian image formation model. Hertzmann and Seitz \cite{HS03} employed some reference objects which are considered to be 
of homogeneous material for simplicity, meaning that purely specular or purely diffuse materials are addressed.
In some other works more complex appearance models are fitted to estimated data, thereby relying
e.g.\ as in the work of Goldman \etal{\hspace{0.2ex}} \cite{{Goldman2005}} on the use of a convex combination of
a small number of known materials, or as in the paper of Oxholm and Nishino \cite{OxNi2014} on 
a probabilistic formulation for linking geometry and lighting estimation by introducing priors.


\begin{figure*}[ht]
\centering
\begin{tabular}{c}
\includegraphics[width=0.9\textwidth ]{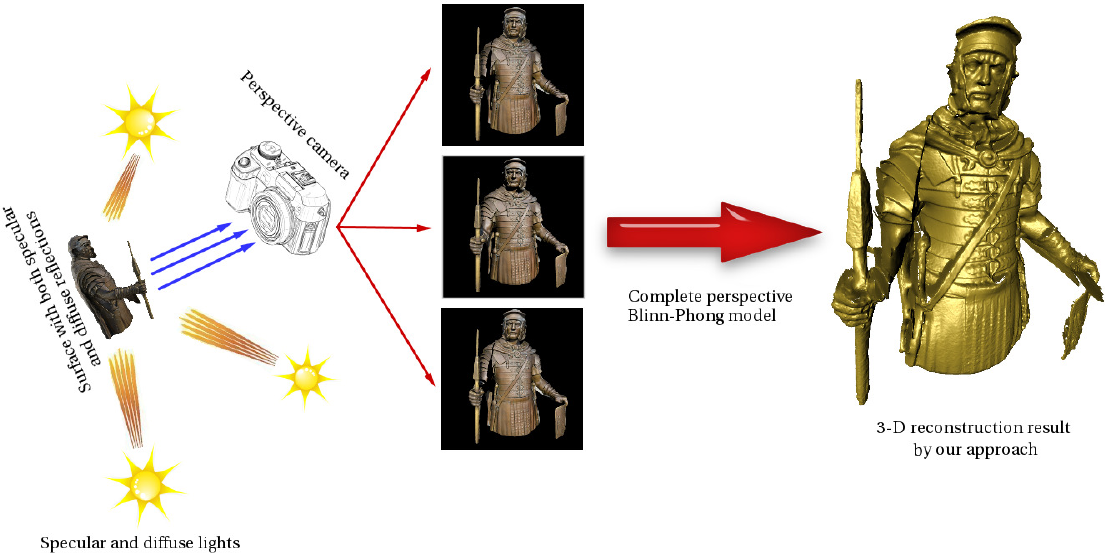}
\end{tabular}
\vspace{1.0ex}
\caption{Highly specular photometric stereo setup illustrated by a complex synthetic experiment. In real-world, surfaces show both specular and diffuse reflections. 
This surface is illuminated by three non-coplanar light sources with both specular and diffuse lights. Shading due to each light is captured in a perspective 
CCD camera. As can be seen, considering all stated assumptions, we are able to recover shape with high degree of surface details.  
\label{FIG1}
}
\end{figure*}

Regarding the perspective projection, one of the first works combining this technique with PS is performed by 
Galo and Tozzi \cite{GT96}. Their work relies on considering point light sources proximate to the lighted object surface. 
A perspective PS model is also proposed by Tankus and Kiryati based on Lambertian reflection \cite{TK05}. A technically 
different perspective method for Lambertian PS using hyperbolic partial differential equations (PDEs) is presented by 
Mecca \etal {\hspace{0.2ex}} \cite{MTB12}.
Turning to the use of non-Lambertian surface reflectance to account for specular highlights in photometric methods, 
we may note that the investigation of a shape-from-shading method using the Phong model has been shown to give very 
reasonable results when employing it within a useful process chain \cite{Vogeletal2009}. Therefore it seems apparent 
that an extension to PS making use of a similar image irradiance
equation may yield even better results given that in PS more input images than in shape from shading are at hand. 

Concerning perspective PS techniques that may also deal with non-Lambertian effects, 
the recent works of Mecca \etal{\hspace{0.2ex}} \cite{MeccaRoCr2015,MeccaQueau2016} should be mentioned. 
In these approaches, an individual model for PS is suggested by considering separated purely specular 
and purely Lambertian reflections using five and ten input images, respectively. 
The separated processing of the reflectance models requires input images with the minimum value of saturation \cite{Tozza2016}
which may lead to cumbersome limitations for some real world applications as e.g.\ in the case
of spatially varying materials. 
When solving the resulting hyperbolic PDEs, Mecca \etal{\hspace{0.2ex}} \cite{MeccaRoCr2015} rely on the 
fast marching method. In order to apply this technique, 
the unknown depth value of a certain surface point must be given in advance. However, this information 
is not always available especially in real-world applications. 
Let us note that a similar approach as in the works of Mecca and his co-authors is also applied in the 
orthographic PS method in \cite{Tozza2016} that is based on 
dividing the surface into two different, purely specular and purely 
diffuse parts, which is a difficult task as also mentioned in \cite{Tozza2016}.

\paragraph*{Our contributions.}
The novel method we propose involves the conceptual advantages of considering perspective projection and 
non-Lambertian reflectance simultaneously based on the complete Blinn-Phong model known from 
computer graphics \cite{Blinn77,phong75}. By taking into account the complete reflection model
our method does not rely on a separation of specular and diffuse reflection in any stage of the computation. 
In particular, no surface or scene dividing task or previous knowledge on the depth 
of the scene is required.
As a side effect our method is inherently able to handle objects with spatially varying materials without modification. 
In addition, it is worth to note that we will use three input images in all our experiments which is the minimum necessary 
inputs for the classic orthographic PS framework with Lambertian 
reflectance model. In recent work it has been discussed that the 
use of three input images can be advantageous \cite{Tozza2016}.

Involving the mentioned model assumptions leads to a concrete PS algorithm as sketched in Fig.\ \ref{FIG1}.
By the combination of the mentioned benefits we propose a more robust and effectively easier to use method
as in previous literature. As a side note, since the complete Blinn-Phong model we employ is extensively studied in 
computer graphics, the surface reflectance in input images as well as expected computational results are potentially 
easier to interprete than in methods that rely on complex preprocessing steps.
Conceptually our work extends the approach presented by Khanian \etal{\hspace{0.2ex}} \cite {khanian15}. 
A main point of the latter conference article is to study the effect of lightening directions 
on numerical stability while the presentation is restricted there to one spatial dimension.

In addition to presenting an appealing alternative PS approach, we investigate two different methods of 
realizing the perspective projection. 
The first method is to compute the normal field and then modifying the gradient field based on the perspective projection 
which is also proposed in 
\cite{Queau15, CVPR2013}. As it manipulates the normal vectors, we refer to this technique as the perspective projection 
based on normal field (PPN) method.
The second method is to consider a perspective parameterization  of photographed object surfaces for getting the 
gradient field of the surface. 
We call this approach the perspective projection based on surface parameterization (PPS) method. 
Furthermore, we investigate the effect of modeling a camera with the charge-coupled device sensor (CCD camera) 
on the reconstruction process and the quality of results.

\section{Perspective Projection}
In this section we introduce two different techniques applied for obtaining the perspective projection. 
As we will also consider for experimental comparison a Lambertian perspective PS model,
we also recall its construction here.
A Lambertian scene with 
albedo $k_d$ is illuminated from directions $L_k=(\alpha _k, \beta _k, \gamma_k)^{\top}$\hspace{-0.1ex}, where $k=1,2,3,$ by corresponding 
point light sources at infinity, with diffuse intensity $l_d$ so that it satisfies the following reflectance equation \cite{BKP1986}:\\
\begin{equation}
\label{eq-rf}
I_k(x,y)
\; = \; 
k_d \left(\frac{L_k \cdot N(x,y)}{{ \|L_k\| }{\|N(x,y)\|}}
\right)l_d 
\end{equation} 
where $k_d$ is the diffuse  material parameter, $I_k(x,y)$ and $ N(x,y)$ are the intensity and surface normal at pixel $ (x,y)$, respectively.


\subsection{Modifying Normal Vectors}

The first perspective projection method deals with processing the field of normal vectors $ N(x,y) ={(n_1(x,y), n_2(x,y),n_3(x,y))}^{\top}$. Once the normal map is reconstructed from the orthographic image irradiance equations, the depth map is recovered by giving the following components in Eq. (2) and Eq. (3) to the integrator:\\
\begin{equation}
p(x,y)
\;
=
\;
\frac{-n_1(x,y)}{ d(x,y)},\hspace{0.1cm}q(x,y)
\;
=
\;
\frac{-n_2(x,y)}{ d(x,y)}
\label{eq-blinnphong}
\end{equation}
where $d(x,y)$ for a camera with the focal length $f$ is:\\
\begin{equation}
d(x,y)
\;
=xn_1(x,y)+yn_2(x,y)+fn_3(x,y)
\label{eq-blinnphong}
\end{equation} 
In what follows, we denote the perspective projection realised via projection of the normal vector with PPN.


\begin{figure}[h]
\vspace{-3.0ex}
\centering
\begin{tabular}{c}{} \\
\includegraphics[width=0.42\textwidth ]{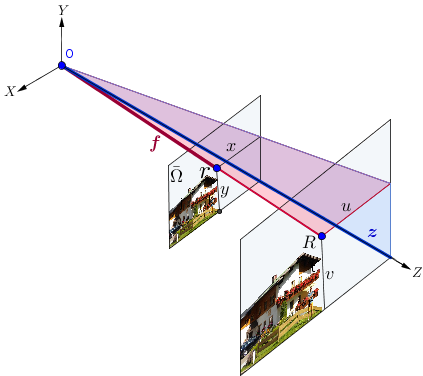}
\end{tabular}
\vspace{0.3ex}
\caption{Perspective projection of the real point $R$ to the image plane $\bar{\Omega} $.  
\label{FIG2}}
\end{figure}


\subsection{Direct Perspective Surface Parameterization }

Another approach to apply the perspective projection is via corresponding surface parameterization, 
shown in Fig. \ref{FIG2}. In order to project the real-world point $R$ to the point $r$ on the image plane $\bar{\Omega}$, we will consider the Thales theorem in both horizontal red and vertical blue triangles. So, we will have:
\begin{equation}
\frac{f}{z(x,y)}
\; = \;
 \frac{x}{u}
\; = \;
 \frac{y}{v}  
\label{eq-sparam}
\end{equation}
On the other hand, in reality the image plane $\bar{\Omega}$ lies behind the lens. 
Therefore, the surface is parameterized using the following formulation, 
where $f$ is the focal length.
\\For all points in $\bar{\Omega}$ as the image plane:\\
\begin{equation}
S(x,y)
\; = \;
\big\{
\begin{array}{rcl}
 \frac{z(x,y)}{f}
\left( -x, -y, f \right)^\top
\, ; \, (x,y) \in \bar{\Omega}  
\end{array}\big\}
\label{eq-sparam}
\end{equation}
\par
From this surface parameterization, we can extract the partial derivatives of the surface:\\
\begin{align}
\hspace{-2ex}S_x= \big(\frac{-z-z_xx}{f}, \frac{-z_xy}{f},  z_x\big)^\top\\
 S_y= \big(  \frac {-z_y x}{f} , \frac{-z_yy-z}{f},   z_y\big)^\top
\end{align}
Finally, we get the surface normal vector $N$ as the cross product of the partial derivatives of the surface:\\
\begin{equation}
N=\frac{z}{f^2} \left( f(z_x,z_y), z_xx+z_yy+z \right)^\top 
\label{eq-normal+}
\end{equation}
So, in this case, the obtained surface normal (8) will be used in image irradiance equation.
We recall here the Lambertian perspective image irradiance equation \cite{TK05}, as this will be extended in our model.

In order to remove the dependency of the image irradiance equation on the unknown depth $z$, it will be substituted
by $\nu=ln (z)$, $z_x=z\nu_x,\;z_y=z\nu_y$, so that we have to apply 
$z=\exp(\nu)$ to obtain the depth $z$ out of our new unknown $\nu$. This yields:
  \begin{equation}
\hspace{-5.7cm}I_k(x,y)\; = \;
\nonumber\\
\end{equation} 
\begin{eqnarray}
k_d 
\frac{f\alpha_k \nu_x+f\beta_k \nu_y+\gamma_k(y\nu_y+x\nu_x+1)^2}
{{\sqrt{(f\nu_x)^2+(f\nu_y)^2+(y\nu_y+x\nu_x+1)^2}}\| L_k \|}
l_d
\label{eq-iirr+}
\end{eqnarray}\\
A closed form solution for the gradient field is obtained in \cite{TK05}. For completeness of the presentation, we now recall the main points in its construction.
Let us consider three input images (the minimum needed inputs in classic PS). By finding $k_d$ from the first image irradiance equation in (9), and replacing it in the second and third image irradiance equation, a linear system of equations $M X=H$ should be solved for obtaining the unknown vector $X=(\nu_x, \nu_y)$:\\
  \begin{equation}
  M=\left( \begin{array}{cc} m_1 & m_2 \\
m_3 & m_4 \end{array} \right)
 \hspace{0.10cm},\;H=\left( \begin{array}{c} h_1 \\ h_2 \end{array} \right) \\ 
\end{equation}\\
where, we have with $ r_i=I_i \| L_i \|$:
\begin{equation}
 m_1= r_2 (f \alpha _1 +x \gamma _1)-r_1 (f \alpha _2+ x \gamma _2)  \\
\end{equation}
\begin{equation}
m_2=r_2(f \beta _1 + y \gamma _1)-r_1 (f \beta _2+ y \gamma _2)
\end{equation}
\begin{equation}
m_3= r_3 (f \alpha _1+ x \gamma _1)-r_1 (f \alpha _3+ x \gamma _3)
\end{equation}
\begin{equation}
m_4=r_3 (f \beta _1 + y \gamma _1)-r_1 f (\beta _3+y \gamma _3)
\end{equation}
\begin{equation}
h_1=-r_2 \gamma _1 +r_1  \gamma _2,\\
h_2=-r_3 \gamma _1 +r_1 \gamma _3
\end{equation}
\label{eq-iirr+}
\\
The explicit solutions are:\\
\begin{equation}
\nu_x
\; = \;
\frac{h_1m_4-m_2h_2}{{m_1m_4-m_2m_3}},
\hspace{0.7ex}\nu_y
\; = \; 
\frac{m_1h_2-h_1m_3}{{m_1m_4-m_2m_3}}
\label{eq-iirr+}
\end{equation}
\noindent
Now, we can obtain the albedo of the surface by plugging the resultant gradient vector for instance into the following equation:\\
\begin{equation}
k_d
\; = \; 
\frac{I_1 \parallel L_1 \parallel \sqrt {(f\nu_x)^2+(f\nu_y)^2+  (y\nu_y+x\nu_x+1)^2}}{l_d{\sqrt{(f\alpha_1\nu_x)+(f\beta_1\nu_y)+\gamma_1(y\nu_y+x\nu_x+1)}}}
\label{eq-iirr+}
\end{equation}

\subsection{Sensitivity of the Solution}
Let us try to access the sensitivity of the solution with respect to the lighting directions, which may lead to conditions
on the illumination. To this end, the non-singularity condition of the matrix of coefficients $M$ introduced in the
previous paragraph should be explored. So, after computing the determinant of $M$ and considering the non-singularity 
condition $det M\neq0$, the non-singularity can be assured in virtually all cases by ensuring that the contributing 
terms are not zero. This idea leads to the indicator: 
\begin{equation}
\left\{ 
\begin{array}{rcl}
\beta_1\alpha_3-\alpha_1\beta_3 
\neq 0\\

 \beta_2\alpha_1-\alpha_2\beta_1 
\neq 0 \\

\alpha_2\beta_3-\beta_2\alpha_3 
\neq 0 \\

y\alpha_1\gamma_1-x \beta_1\gamma_1
\neq 0\\

x\beta_2\alpha_1-y\alpha_2\gamma_1 
\neq 0\\

y\alpha_2 \gamma_3-x\beta_2\gamma_3 
\neq 0\\

x\gamma_1\beta_1-y\gamma_1\alpha_1 
\neq 0\\

y\gamma_1\alpha_3-x\gamma_1\beta_3 
\neq 0\\

y\gamma_2\alpha_1-x\gamma_2\beta_1 
\neq 0\\

y\alpha_2\gamma_1-x\beta_2\gamma_1 
\neq 0\\

x\gamma_2\beta_3-y\gamma_2\alpha_3 
\neq 0\\
\end{array}\right\}
\label{eq-sparam}
\end{equation}
The first three expressions imply the linear independence of light directions and it can be also obtained from the non-singularity
condition of the light directions matrix. The other resultant expressions are different and satisfying all 
of them may not be an easy task. Consequently, the sensitivity of the solution to the lightening can be higher than 
in the PPN approach.
\begin{figure*}[!ht]
\centering
\begin{tabular}{ccc}
\hspace{-5mm}
\includegraphics[width=0.3\textwidth ]{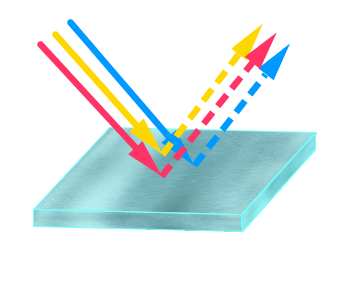}
\vspace{-2mm}
&
\hspace{-14mm}
\includegraphics[width=0.3\textwidth]{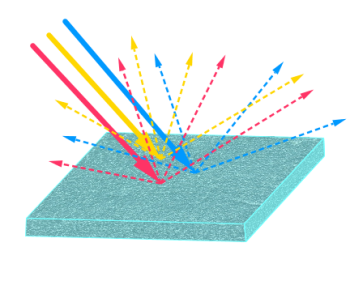}
&
\hspace{-4mm}
\includegraphics[width=0.35\textwidth]{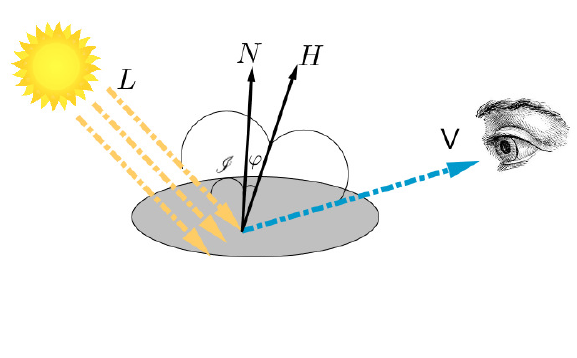}
\\
\hspace{-18mm}
Specular reflection
&
\hspace{-22mm}
Diffuse reflection
&
\hspace{-8mm}
Blinn--Phong\ model
\end{tabular}
\vspace{1.0ex}
\caption{ Left and middle: Specular and diffuse reflections. Both reflections can exist simultaneously in different parts of a real-world object.   
Right: Vectors and angles applied in the Blinn-Phong model. 
\label{FIG3}
}
\end{figure*}
\section{Blinn-Phong Reflectance Model} 

Let us introduce the Blinn-Phong reflectance model for addressing the issue of specular reflections 
of non-Lambertian materials. A useful reflectance model giving an approximation for real-world surface reflectance is considering additionally to the Lambertian reflectance a specular term as introduced in the model of Blinn-Phong~\cite{Blinn77,phong75}. We stress the world "additionally" since in reality, most of the objects show both of these reflections in different areas, cf. Fig. 3. Therefore, they include both reflection models at the same time. In the Blinn-Phong model, angle of incidence  $\mathcal{I}$ and also the angle $  \varphi$ between the vector $N$ and the vector $H$ (halfway vector of the light and viewing direction) are applied as shown in Fig. 3. Now we consider the Blinn-Phong model under the  perspective projection. To this end, we apply again the two different mentioned perspective approaches. 
The basic and complete Blinn-Phong image irradiance equation is defined as:
\\
\begin{equation}
\hspace{-3cm}I(x,y)
\;
=
\;
k_d
\left(
\frac{L\cdot N(x,y)}{ \|{ L}\|  \|{N(x,y)\|}}
\right)
l_d
\nonumber\\
\end{equation}
\begin{eqnarray}
+
k_s \left(
\frac{H(x,y,z)\cdot N(x,y)}{\|{H(x,y,z)\|}\|{ N(x,y)}\|}\right)^n 
l_s
\label{eq-blinnphong}
\end{eqnarray}
\noindent
Here $k_s$ is the specular material parameter.  
$l_s$ is the specular light source intensity and the exponent $n$ is also called the specular sharpness or shininess. 
\par
A corresponding orthographic model has been investigated in the shape-from-shading context
in \cite{camillitozza2017}. To develop the perspective Blinn-Phong PS model, we focus on the surface parameterization  and plug in the perspective normal \eqref{eq-normal+} in \eqref{eq-blinnphong}.

Considering $k$ input images for
corresponding lighting directions, this yields after some computation
the perspective  Blinn-Phong reflectance equations as:
\\
\begin{equation}
{I_k(x,y)
=
k_d l_d 
\frac{f\nu_x \alpha_k+f\nu_y \beta_k+
       \gamma_k w}
{{g\sqrt{(f\nu_x)^2+(f\nu_y)^2+(w)^2}}}}+ k_s l_s
\nonumber\\
\end{equation}
\begin{eqnarray}
\left( 
\frac{
f\nu_x \alpha_kp-xg
+f\nu_y \beta_kp-yg
+
wd
}
{
\sqrt{ r +(w)^2} p \|{
D\| }
}
\right)^{n}
\label{eq-finalblinnphong}
\end{eqnarray}\\
where 
\begin{equation} 
\hspace{-1cm}p:= \; \|{ (x,y,f)\|},\quad g:= \|{ L_k}\|
\end{equation}
\begin{equation}
r : = f^2(\nu_x^2+\nu_y^2),\quad d:= (\gamma_k P-f\|{L_k}\|)
\end{equation}
\begin{equation}
D:=
\left[
\begin{array}{c}
\alpha_k- G  x\\ \beta_k- G y\\ \gamma_k- Gf 
 \end{array}
\right]
\end{equation}
\begin{equation}
w=(y\nu_y+x\nu_x+1)
\end{equation}

\subsection{Numerical approach}

Now, we present the numerical procedure which can be applied for addressing such a highly nonlinear system of equations. 
Recalling the description of a system of equations as $F(X)=0$, where $F:{\mathbb{R}}^n \longrightarrow {\mathbb{R}}^m$  is a given function by the equations from (20), we will discuss our solution procedure.

In order to cope with such a nonlinear system of equations, we applied the Levenberg-Marquardt method introduced in \cite{leven44,Marq63} as a combination of the Gau\ss-Newton method and steepest descent direction technique. In this method, if $ X^k$ is the point at iteration $k$, the next iteration can be computed as:\\
 \begin{equation}
X^{k+1}:=X^{k}+{d}^k,
\label{eq-def-ij}
\end{equation}
\noindent
\begin{equation}
d^{k}:=-(JF(X^{k})JF(X^k)^{T}+ \lambda_k I)^{-1} JF(X^{k})F(X^{k})
\label{eq-iirr+}
\end{equation}
 with $ \lambda _k> 0$.

The matrix $ JF(X^{k}) JF(X^k)^{T}+\lambda_k I$ is positive definite and $ d^{k}$ is well-defined. In addition, this method does not need the conditions such as the invertibility of Jacobian matrix or Hessian matrix  or $ m=n$.

Our numerical approach for the PPS method is based on the following problem formulation. Recalling the perspective Blinn-Phong reflectance equations \eqref{eq-finalblinnphong}, and dividing three  equations ($I_1/I_2$, $I_2/I_3$, $I_1/I_3$, corresponding to the three used images in our method) leads to a non-linear system of equations, with the equations like the following equation \eqref{eq-def-devide} as obtained for dividing the $M^{th}$ and  $N^{th}$ images:
\begin{equation*}\begin{split}
I_M(x,y){\parallel L_M\parallel}
\bigg(\alpha_Nf\nu_x+\beta_Nf\nu_y+\gamma_Nw\bigg){{\parallel Q\parallel}}^{n}
\end{split}\end{equation*}

\begin{equation*}\begin{split}
 \bigg({(f\nu_x\alpha_N+f\nu_y\beta_N)(-P)-(x+y)({\parallel L_N\parallel})+wk}\bigg)^{n}
\end{split}
\end{equation*}

\begin{equation*}\begin{split}
-I_N(x,y){\parallel L_N\parallel}
\bigg(\alpha_Mf\nu_x+\beta_Mf\nu_y+\gamma_Mw\bigg){{\parallel T\parallel}}^{n}
\end{split}
\end{equation*}

\begin{equation}\begin{split}
\bigg({(f\nu_x\alpha_M+f\nu_y\beta_M)(-P)-(x+y)({\parallel L_M\parallel})+we}\bigg)^{n}=0
\end{split}
\label{eq-def-devide}
\end{equation}
with\\
\begin{equation}
Q:=
\left[ 
\begin{array}{c}
\alpha_M-{\parallel L_M\parallel}x\\
\beta_M-{\parallel L_M\parallel}y\\
\gamma_M-{\parallel L_M\parallel}f
 \end{array}
\right]
\hspace{0.2ex},
\hspace{0.4ex}
T:=
\left[ 
\begin{array}{c}
\alpha_N-{\parallel L_N\parallel}x\\
\beta_N-{\parallel L_N\parallel}y\\
\gamma_N-{\parallel L_N\parallel}f
 \end{array}
\right]
\end{equation}

\begin{equation}
k=(\gamma_NP-f\parallel L_N\parallel),\quad e=(\gamma_MP-f\parallel L_M\parallel)
\end{equation}

\noindent
It should be noted that even in this case of existing specularities and 
in the process of solving the perspective PS system for the Blinn-Phong model \eqref{eq-finalblinnphong}, 
we will still follow Woodham and make use of only three input images.

Furthermore, as for the case of Lambertian PS, we will also deal with the Blinn-Phong model 
using the perspective version based on transforming the normal vectors (PPN method), i.e.\
after orthographic Blinn-Phong PS. Finally, the obtained gradient fields are 
processed by the Poisson integrator, see e.g.\ \cite{baehretal2016} for a recent account of surface normal integration. \\
\begin{figure*}[!ht]
\centering
\begin{tabular}{ccc}
\hspace{-4mm}
\includegraphics[width=0.32\textwidth ]{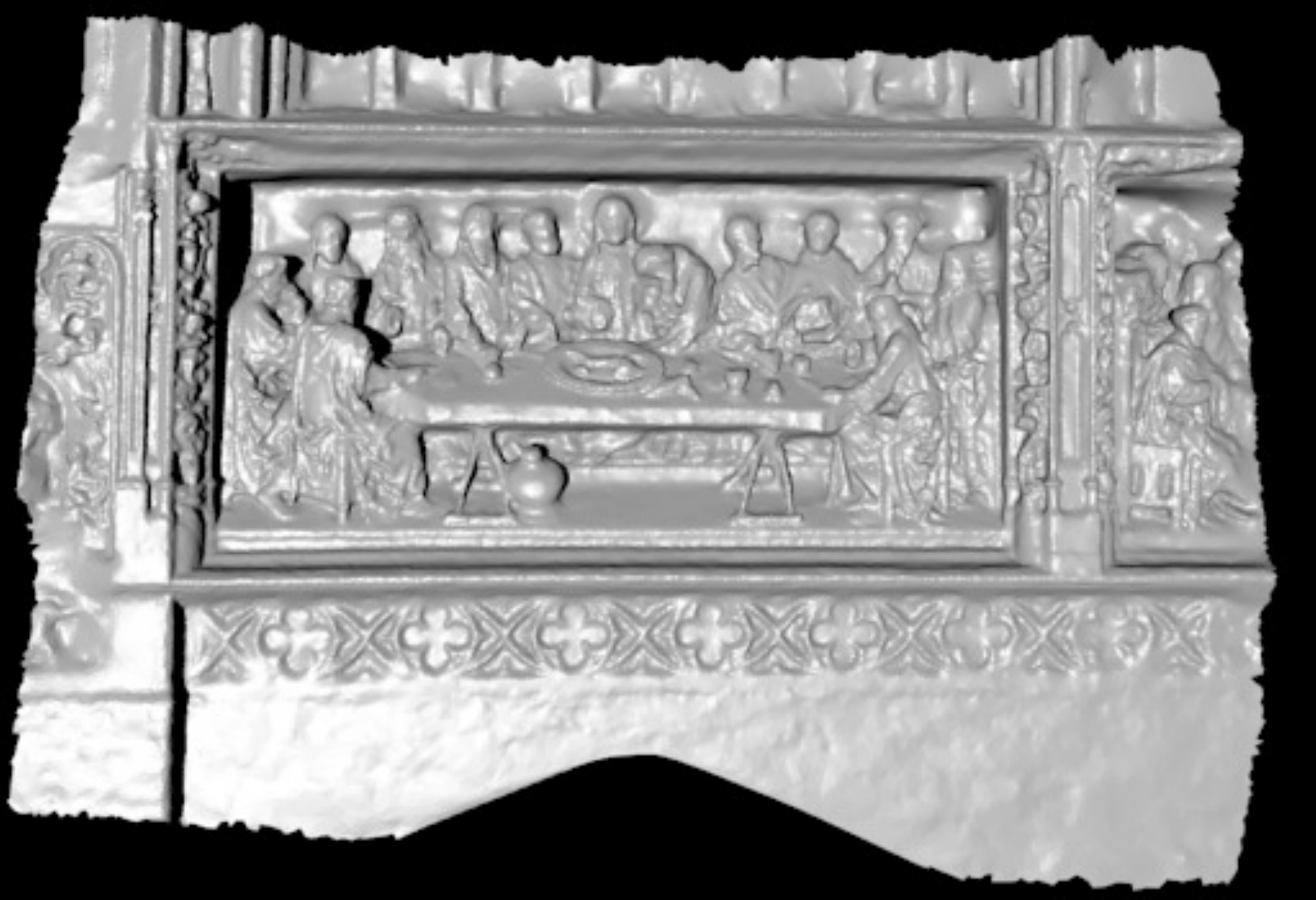}
&
\hspace{-2mm}
\includegraphics[width=0.32\textwidth]{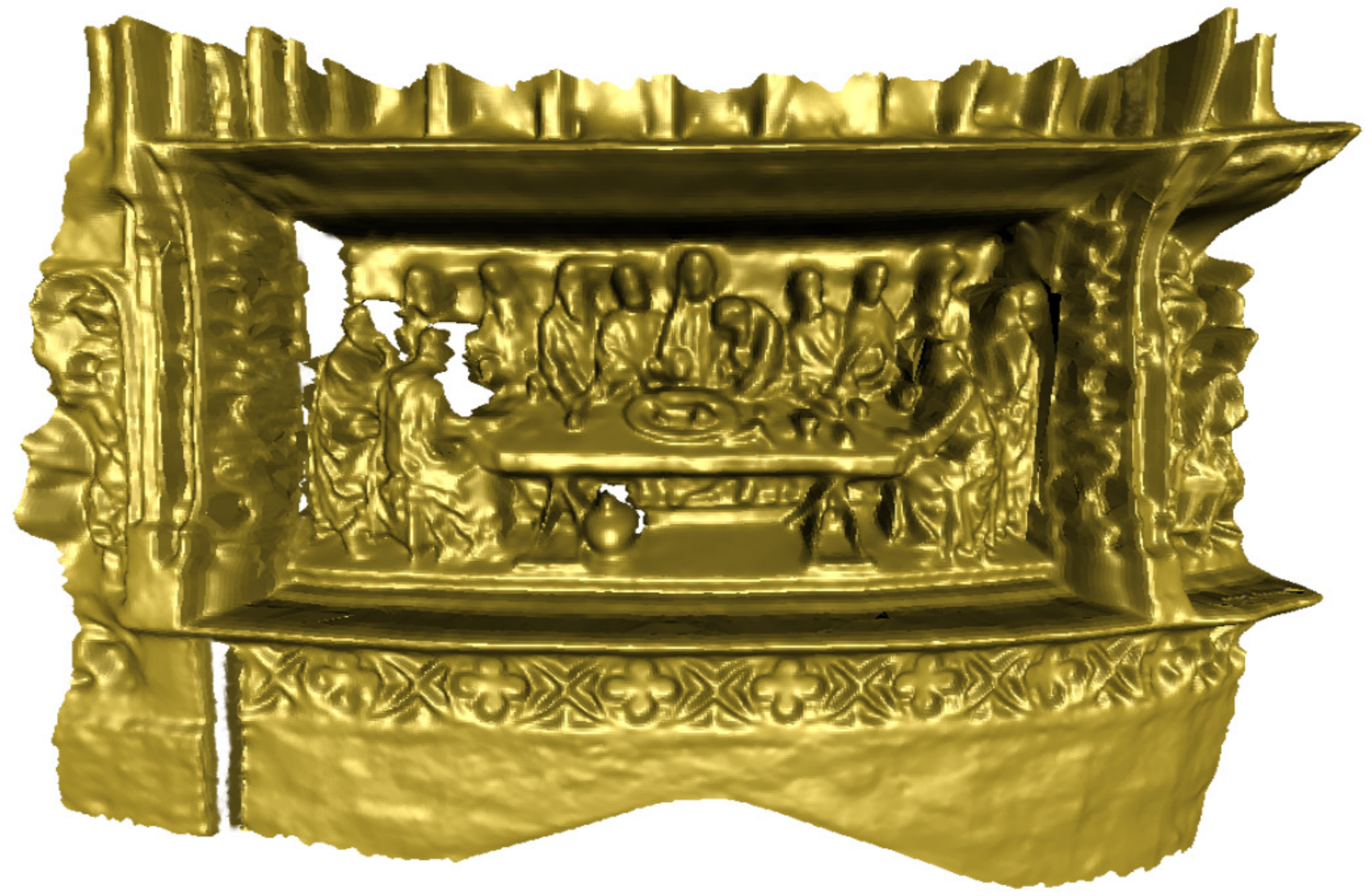}
&
\hspace{-2mm}
\includegraphics[width=0.32\textwidth]{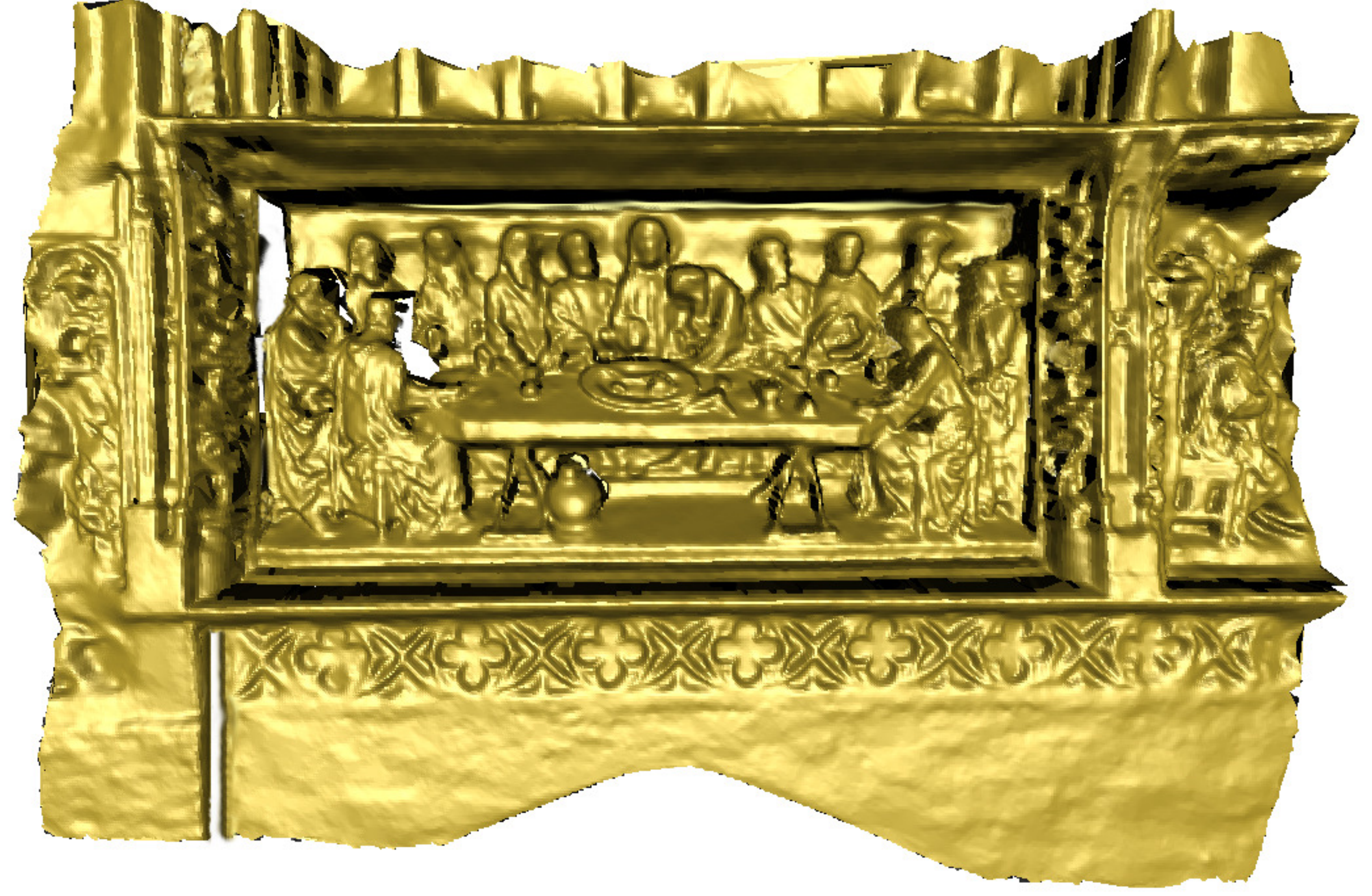}
\end{tabular}
\vspace{1.0ex}
\caption{ Comparison of the surface reconstruction techniques. Left: Input image. Middle: Our 3-D reconstruction using orthographic projection. Right: Our 3-D reconstruction by perspective projection. It can be observed that the perspective approach is able to generate a more compatible result with respect to the original image.
\label{FIG4}
}
\end{figure*}
\begin{figure*}[!hb]
\centering
\begin{tabular}{ c}
\includegraphics[width=1\textwidth ]{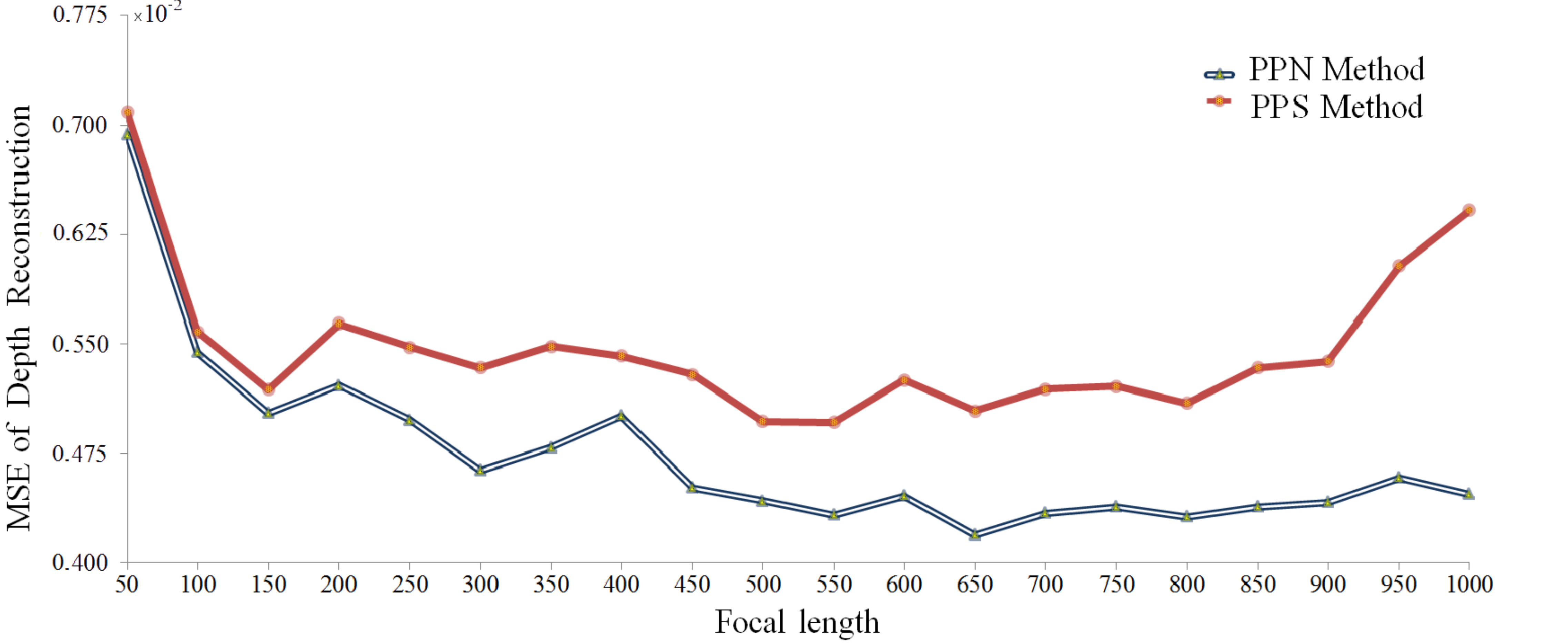}
\end{tabular}
\vspace{-1.0ex}
\caption{Comparing two described perspective methods regarding their depth reconstructions. 
\label{FIG5}
}
\end{figure*}
\section{CCD Cameras} 
We will also investigate the modeling of the CCD camera. In the case of CCD cameras, the following projection mapping is used as presented in \cite{Hartley2004}.
The matrix\\
\begin{equation}
 \Gamma=
 \left[ \begin{array}{ccc}
 	\psi_x & \xi & \delta_x
\\ 0 & \psi_y & \delta_y 
\\ 0 & 0& 1 
\end{array} \right]
\end{equation}
contains the intrinsic parameters of the camera, namely the focal length in $x-$ and $y-$ direction 
equal to $ \psi_x=\frac{f}{h_x}$ and $ \psi_y=\frac{f}{h_y}$, with the sensor sizes $ h_x$ and $h_y$   
and the principal point or focal point $(\delta_x, \delta_y)^\top$. The parameter $\xi$ is called skew parameter. Here, we neglect this parameter since it will be zero for most of normal cameras \cite{Hartley2004}.  Using this matrix, we will introduce the 
following transformation to convert the dimensionless pixel coordinate $X=(x,y,1)^\top$ to the
image coordinate $\chi=(c,d,f)^\top$ as follows:\\
\begin{equation}
X=\Gamma\hspace{0.6ex} {\frac{1}{f}}\hspace{0.6ex} \chi \\
\end{equation}
By applying the above-mentioned transformation, the following representation for the projected point $\chi$ will be obtained: \\
\begin{equation}
\left[ \begin{array}{c}
c  
\\ d 
\end{array} \right]=
 \left[ \begin{array}{c}
h_x(x-\delta_x)  
\\ h_y(y-\delta_y)   
\end{array} \right]
\end{equation}
The effect of this modeling can be potentially interesting, since this information is not always accessible.
The above transformation is called \emph{centerizing} in the experiments.



\begin{figure*}[!ht]
\centering
\begin{tabular}{ c }
\hspace{-1ex}\includegraphics[width=0.8\textwidth ]{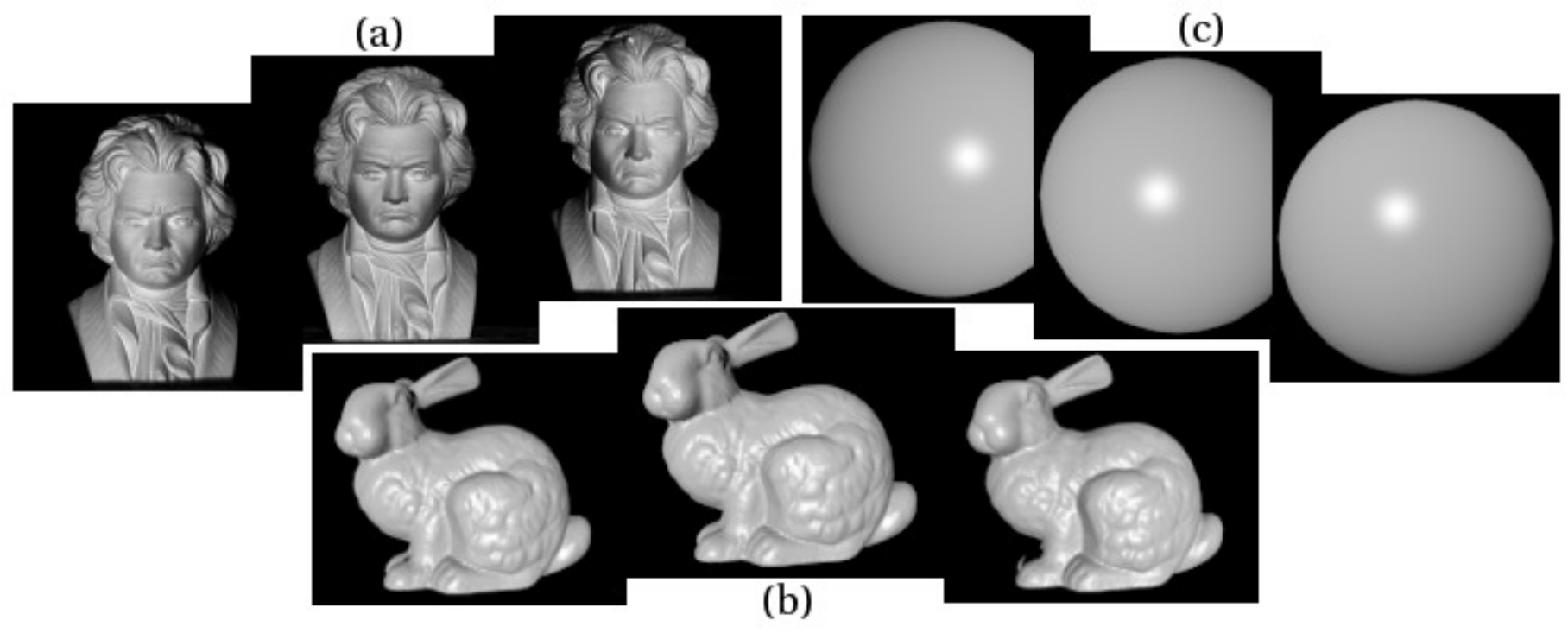}

\end{tabular}
\vspace{-1.0ex}
\caption{Set of three test images used for our 3-D reconstruction:
(a) Real scene used for reprojecting; (b) and (c) are rendered images used for our 3-D reconstruction in presence of specularity. 
\label{FIG6}}
\end{figure*}

\begin{figure*}[!hb]
\centering
\begin{tabular}{ ccc}
\includegraphics[width=0.2\textwidth ]{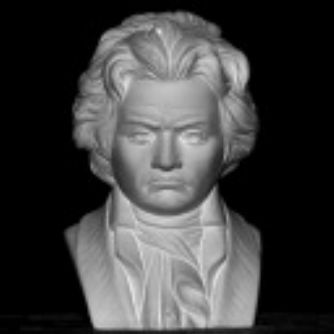}
&
\includegraphics[width=0.2\textwidth] {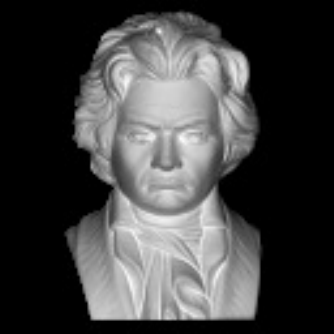}
&
\includegraphics[width=0.2\textwidth] {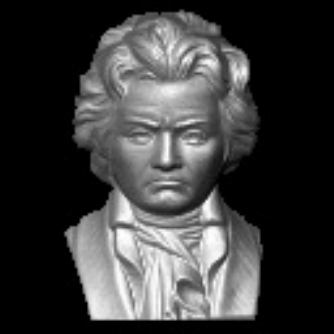}
\end{tabular}
\vspace{1.0ex}
\caption{An account of reprojected Beethoven images. Left: Second input image for PS. 
Middle: Reprojected second image obtained from PPN method. 
Right: Reprojected second image using the PPS technique. 
\label{FIG7}
}
\end{figure*}
\begin{table*}[!hb]
\caption{Comparison between MSE of the reprojected Beethoven images from two described perspective methods of PPN and PPS. \label{table-1}}
\vspace{-5mm}
\begin{center}
\begin{tabular}{ | l | c | c | c | }
  \hline
  Perspective method & MSE for 1st input & MSE for 2nd input  & MSE for 3rd input \\ \hline \hline
   PPN method & 0.004239 & 0.003297 & 0.007535 \\ \hline
  PPS method & 0.008042 & 0.021409 & 0.007644 \\ \hline
\end{tabular}\vspace{1.0ex}
\end{center}
\end{table*}
\section{Experiments} 
This section describes our experiments performed by the proposed model and approaches. 
In a first test we confirm the investigation of Tankus \etal \, \cite{TSY2005}
that the use of an orthographic camera model may yield apparent distortions in the reconstruction
while a perspective model may take the geometry better into account, see the experiment documented in Fig. \ref{FIG4}. 
This justifies the use of the perspective camera model. Note that in the figure the object of interest
is relatively close to the camera.

In a series of tests we now turn to quantitative evaluations of the proposed computational models.
To this end we consider the set of test images for use in the next experiments as shown in Fig. \ref{FIG6}. The Beethoven test images 
(which depict a real-world scene) and the Sphere images are of the size $128\times 128$. The Stanford Bunny 
test images have a resolution of $150\times 120$. Both Bunny and Sphere are rendered using Blender. The 3-D model of Stanford Bunny is obtained from the Stanford 3-D scanning repository \cite{Bunny004}.
The 3-D model of the face presented in Fig. \ref{FIG9}  is taken from \cite{Face} with the size of $256\times 256$. For comparing our results, the ground truth depth maps are extracted, and we will make use of the Mean Squared Error (MSE) showing the accuracy.

After considering the mentioned test settings, we demonstrate the applicability of our method at hand of real world medical
test images from gastro endoscopy and discuss its superior reconstruction capabilities compared to previous models.
\subsection{Tests of accuracy}
In the first evaluation, we compare results of two mentioned perspective techniques of PPN and PPS, applied on the specular Sphere in Fig. \ref{FIG6} (c) with different values of focal length.
MSE results of these 3-D reconstructions are shown in Fig. \ref{FIG5}. While obtained results of described perspective methods for some low values of focal lengths are close to each other, PPN perspective strategy outperforms PPS as the focal length increases.

In the second experiment concerned with the Beethoven image set, 
we investigate the difference between two mentioned perspective approaches on a more complex real-world object scene.
To this end, we give in Table \ref{table-1} the MSE comparing 
gray value data of the reprojected and input images. Since in this case the ground truth depth map is not available, we reconstruct the reprojected images by obtaining the gradient fields from the mentioned perspective approaches and replacing them in the Lambertian reflectance equation.
It can be deduced from Table \ref{table-1} that reprojecting from PPS method reaches a close 
accuracy regarding the third input image, while the PPN approach achieves higher 
accuracy in terms of the first and especially second input image.

As the reprojected images in Fig. \ref{FIG7} show, the difference between these methods as given in Table \ref{table-1} can be quite significant. Furthermore, it is indicative of higher sensitivity of the PPS method to the lightening than the PPN approach.

\subsection{Perspective methods and CCD camera model}
\begin{figure*}[!ht]
\begin{tabular}{c}
\hspace{20mm}\includegraphics[width=0.7\textwidth]{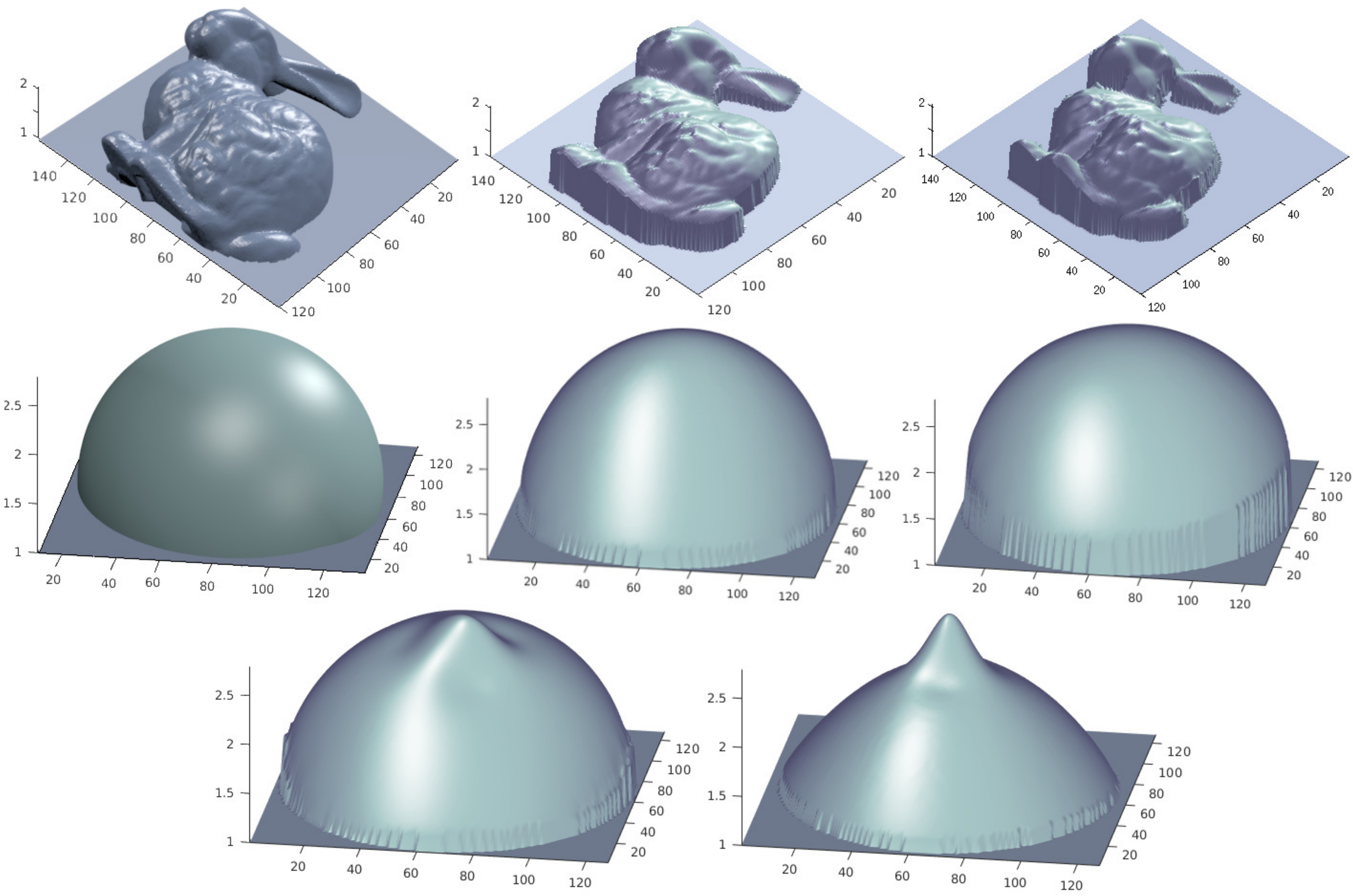}
\end{tabular}
\caption{First and second row: Left: Groud truth. Middle: depth reconstruction from complete Blinn-Phong model with PPN approach. Right: depth reconstruction from complete Blinn-Phong model with PPS approach. These results turn out the proficiency of the proposed method for appealing reconstruction of the images including strong specularities. In addition, PPN approach achieves more faithful reconstructions. Last row: Depth reconstruction from Lambertian model in the presence of specularity accompanied by different perspective projection. Left: PPN approach. Right: PPS method. As it can be seen, the Lambertian model is not able to provide a faithful reconstruction for the specular surface.
\label{FIG8}
}
\end{figure*}

Table \ref{table-2} and Fig. \ref{FIG8} present the results of our 3-D reconstructions for highly specular input images as shown in Fig. \ref{FIG6} (b) and Fig. \ref{FIG6} (c), respectively. In order to produce such images, we set non-zero intensities for diffuse and also specular light. Furthermore, the objects include both diffuse ans specular reflections.
\begin{table*}[!ht]
\caption{MSE of the reconstructed depth from images with specularities by two perspective methods of PPN and PPS. As it is clear, we consider 3-D reconstruction in the presence of both diffuse and specular reflection simultaneously from the surface which leads to involving both $k_d$ and $k_s$ and applying complete Blinn-Phong model. In addition, we applied both diffuse and specular light. Finally, we extended our model to different perspective projection techniques.    
\label{table-2}}
\begin{center}
\begin{tabular}{ | l | c | c |c | c | c |c | c |c | }
  \hline
Reconstruction by Perspective Blinn-Phong PS& $k_d$ &$k_s$& $l_d$ &$l_s$ & shininess   &centerizing & no centerizing \\ \hline \hline
  MSE of PPN method for  Bunny & 0.6&0.4 & 1.2 & 1.2 &  50 & 0.006355  & 0.042082  \\ \hline
  MSE of PPS method for  Bunny & 0.6&0.4 & 1.2 & 1.2 &  50 & 0.012318  & 0.011318  \\ \hline \hline
  MSE of PPN method for Sphere & 0.5&0.5 & 1.2 & 1.2 & 150 & 0.008264  & 0.022568  \\ \hline
  MSE of PPS method for Sphere & 0.5&0.5 & 1.2 & 1.2 & 150 & 0.008431  & 0.007716  \\ \hline
\end{tabular}\vspace{-2.0ex}
\end{center}
\end{table*}
\begin{center}
\begin{figure*}[!ht]
\begin{tabular}{c}
\hspace{15mm}\includegraphics[width=0.85\textwidth]{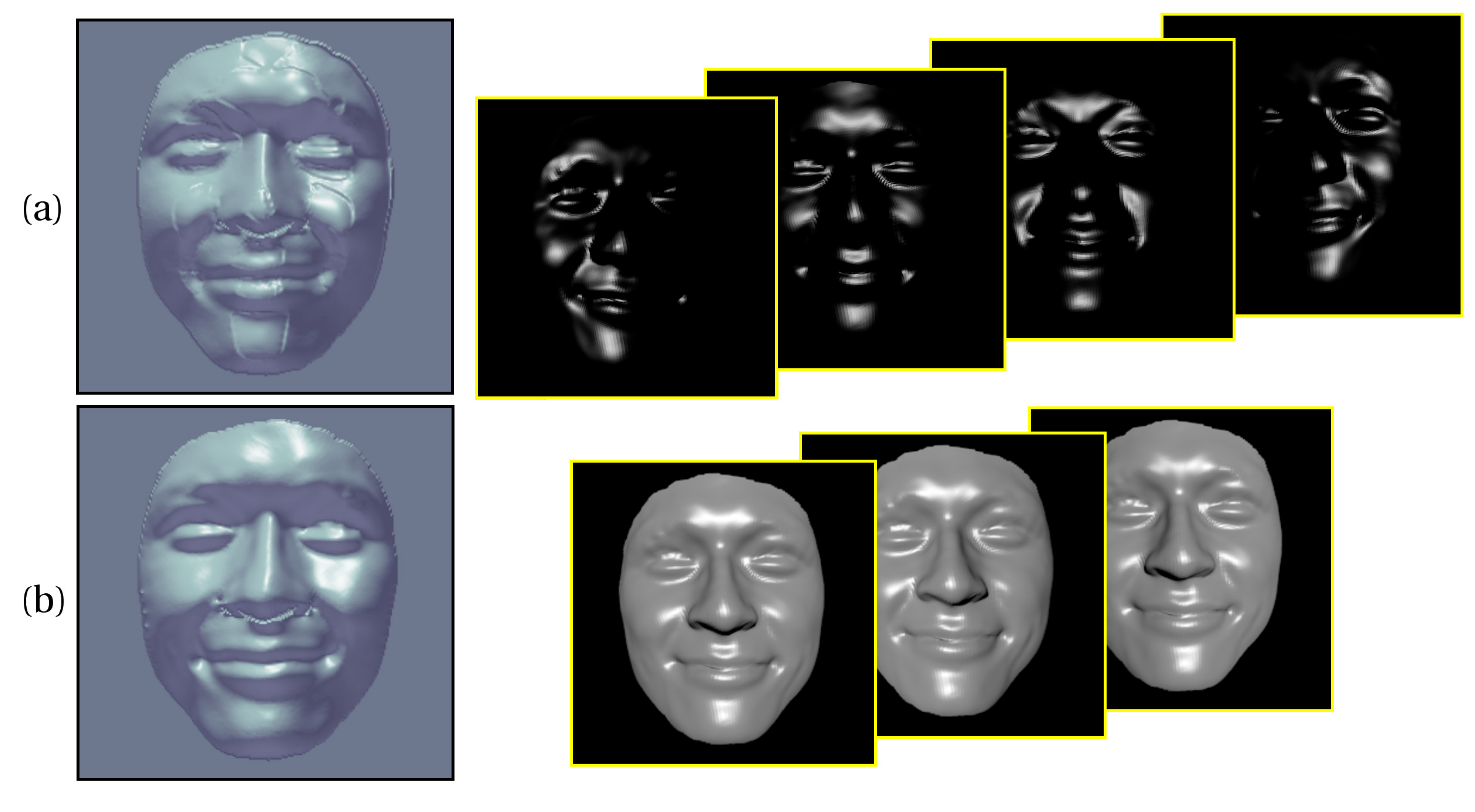}
\end{tabular}
\caption{First row: Four purely specular input images as applied in purely specular model of \cite{MeccaRoCr2015} and the 3-D reconstruction of the \cite{MeccaRoCr2015} approach which shows deviations especially around the highly specular areas. Second row: Three ordinary input images including both diffuse and specular components as the input of our method and Our 3-D reconstruction. Note that our method does not need the decomposition of the input images into purely diffuse and purely specular components which is a very difficult task even for synthetic images.  
\label{FIG9}
}
\end{figure*}
\end{center}
\begin{table*}
\caption{MSE of the reconstructed depth from images with high specularities shown in Figure \ref{FIG9}.
\label{table-3}}
\begin{center}
\begin{tabular}{ | c | c | c | c | c | c | c | }
\hline
Depth reconstruction approach & $k_d$ & $k_s$ & $l_d$ & $l_s$  &shininess & MSE
\\ 
\hline\hline
Proposed method              & 0.3 & 0.7 & 1.2 & 1.2 & 50&\textbf{0.004019}  \\  \hline
Mecca \etal{\vspace{0.2ex}}\cite{MeccaRoCr2015}& 0 & 0.7 & 0 & 1.2 &50&  0.056586   \\ \hline 
\end{tabular}\vspace{-2.0ex}
\end{center}
\end{table*}
\begin{figure*}[h]
\centering{
\begin{tabular}{c}
\includegraphics[width=1\textwidth]{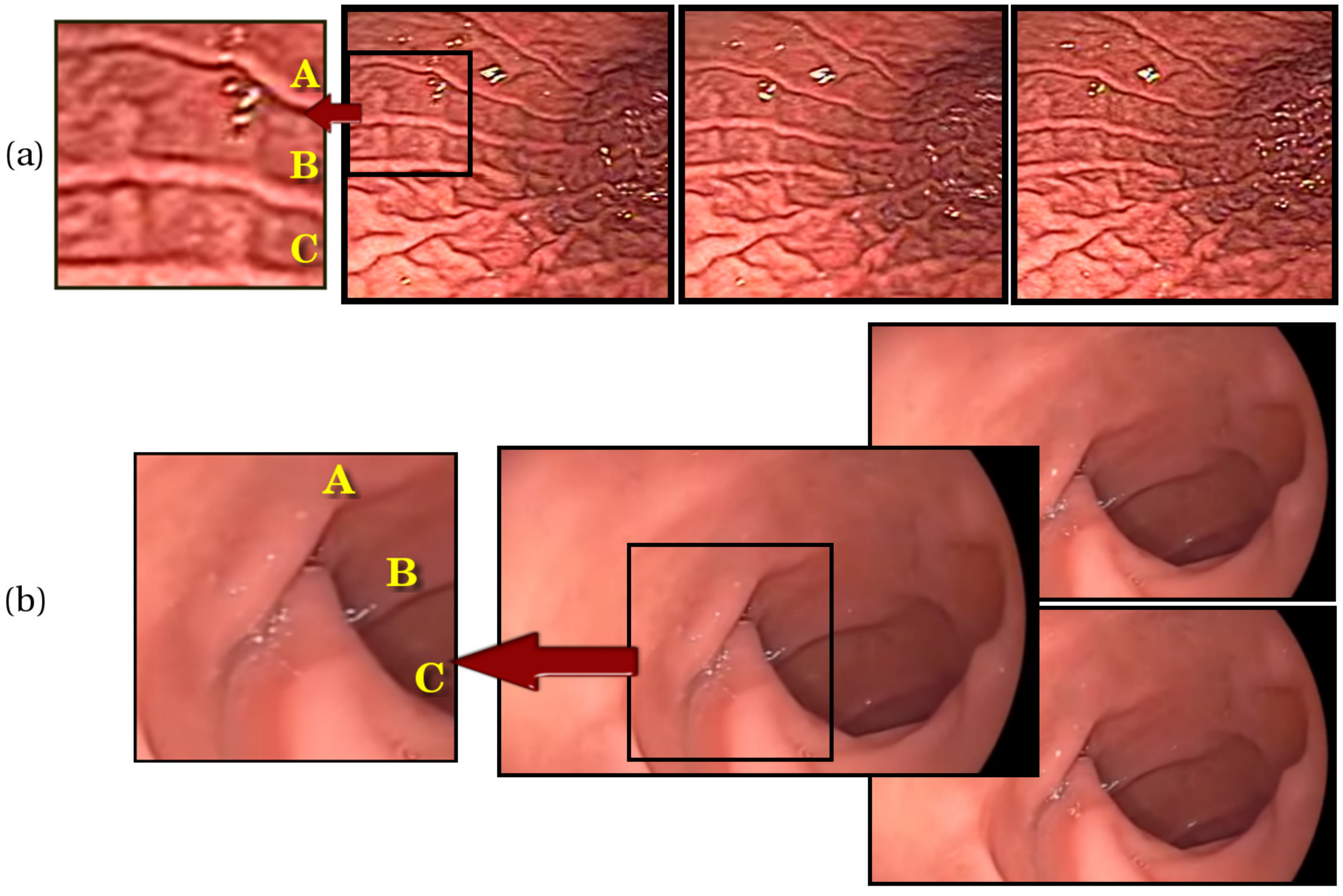}
\\
\end{tabular}
\caption{Test images with high specularity used in realistic real world senario. These images are produced in an endoscopy experiment. So, they do not benefit from any laboratory facilities or confine to the controlled setup conditions.
\label{FIG10}
}}
\end{figure*}
Values of MSE for 3-D reconstruction show the high accuracy of our depth reconstructions by applying the complete Blinn-Phong model which is accompanied by two presented perspective schemes. On the other hand, while results of the recovered depth map for the sphere are close to some extent, the outcome of the computed depth map for Bunny based on the PPN method obtains higher accuracy. However, the table also illustrates the higher sensitivity of the PPN perspective scheme to centerizing transformation than the PPS perspective method.

Finally, we compare our approach with the Lambertian model which is the most common model applied in PS and also the method presented by Mecca \etal{\hspace{0.1ex}} \cite{MeccaRoCr2015}.
Fig. \ref{FIG8} shows the outcome of applying the Lambertian model. The deviation from faithful reconstruction over the specular area of the surface can be seen clearly. 

The comparison between our approach and \cite{MeccaRoCr2015} is also shown in Fig. \ref{FIG9}. 
As already indicated, our method applies complete perspective Blinn-Phong model on three images including both diffuse and specular reflections 
and lights, while the method in \cite{MeccaRoCr2015} uses the specular term in Blinn-Phong model to handle four purely specular images. 
The excellent result of the proposed method presented in Fig. \ref{FIG9} (b) over the high value of specularity with the absence 
of any deviation or artifact shows that the proposed method outperforms state-of-the-art approaches such as in \cite{MeccaRoCr2015}. 
The MSE values of 3-D reconstruction associated with experiments in Fig. \ref{FIG9} are 
also illustrated in Table \ref{table-3}.

\subsection{Tests of applicability on real-world test images}\label{Tests of applicability:}

This section describes experiments conducted by the proposed approach on realistic images. 
Let us first turn to some real-world medical test images. It should be noticed that we may also call these images realistic because we did not benefit from a controlled setup or additional laboratory facilities. We used just the images that are available as in any kind of medical (or many other real world) experiments. Let us note that experiments with endoscopic images are well known to yield a challenging test for photometric methods, and they are widely accepted for indicating possible medical applications of photometric approaches, see e.g.\ \cite{TSY-PR2004,Tatemasu2013}.
As for our work, the usefulness of computational results for the indicated, concrete medical application is confirmed via collaboration with specialized medical doctors.\footnote{Let us mention as a reference the collaboration with 
Dr.\ Mohammad Karami H. (Dr.mokaho@skums.ac.ir) who is a gastroenterologist and internal medicine specialist at Shahrekord University of Medical Science (Iran).}
\par
\begin{figure*}[!h]
\centering{
\begin{tabular}{ c }
\hspace{-10mm}
\includegraphics[width=0.75\textwidth]{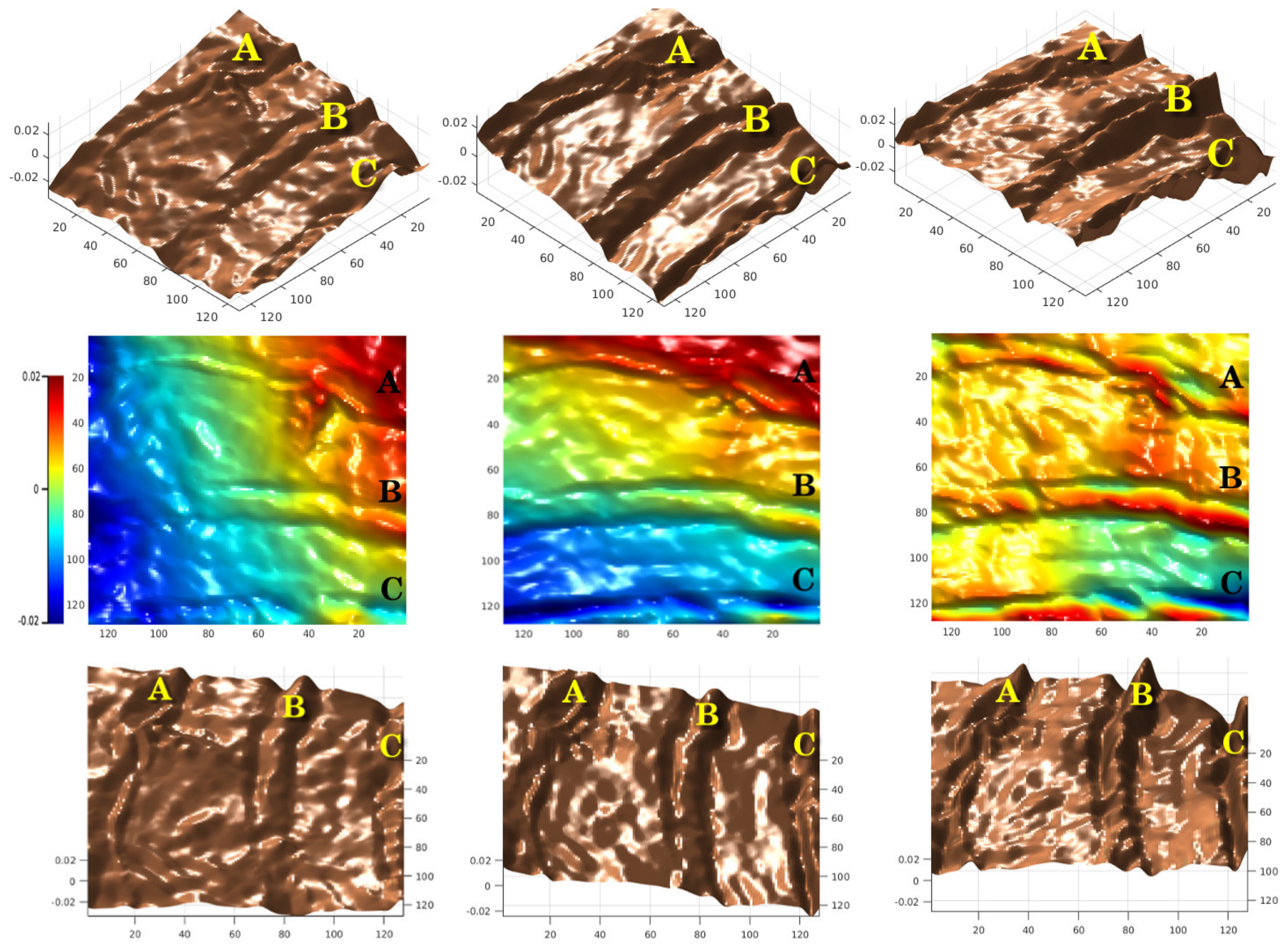}
\end{tabular}
\caption{Depth reconstruction from real world endoscopy images:
(first column) results of Lambertian model, (second column) results of the first proposed method (complete Blinn-Phong using PPS) and (third column) results of second proposed approach (complete Blinn-Phong model using PPN). The deviation in the Lambertian results can be clearly seen, while the results of our approach provide faithful 3-D reconstruction without any deviation and also with a high amount of details.} 
\label{FIG11}}
\end{figure*}
\begin{figure*}[!h]
\centering
\begin{tabular}{c}
\hspace{-5mm}
\includegraphics[width=0.75\textwidth]{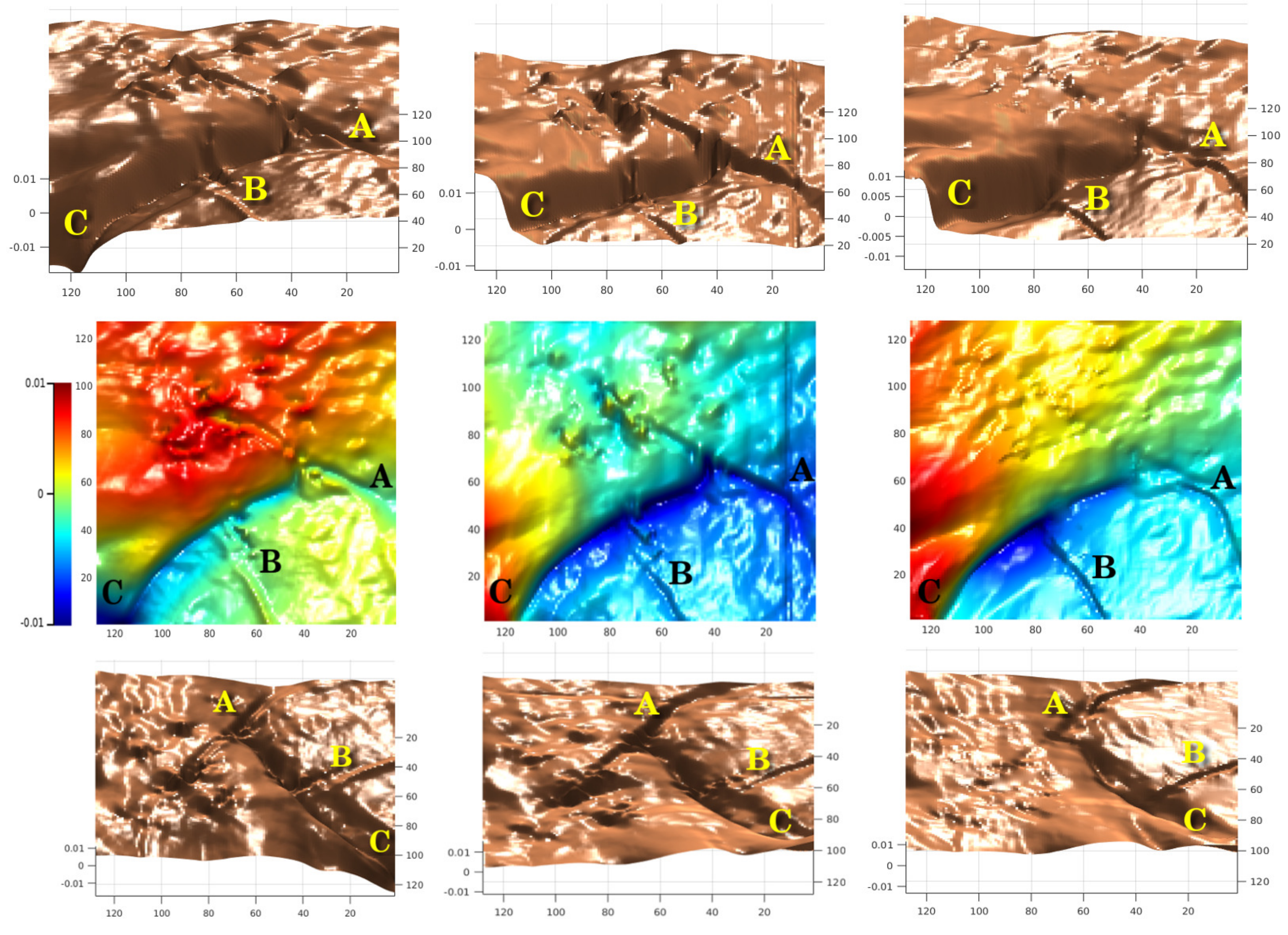}
\end{tabular}
\caption{Depth reconstruction from real world endoscopy images:
(first column) results of Lambertian model, (second column) results of the first proposed method (complete Blinn-Phong using PPS) and (third column) results of second proposed approach (complete Blinn-Phong model using PPN). Once more, the deviation in Lambertian outcomes is clear, whereas our approach provides a trustable 3-D reconstruction without any deviation.
\label{FIG12}
}
\end{figure*}

We have performed trials on endoscopic images in which existence of high specularities is unavoidable. 
Input images are presented in Fig. \ref{FIG10} (a) and Fig. \ref{FIG10} (b) which are endoscopies of the upper gastrointestinal system. Their 3-D reconstructions are represented in Fig. \ref{FIG11} and Fig. \ref{FIG12}.
\par
Similar to all the previous experiments only three input images are used. All outputs are displayed with an identical viewpoint enabling their visual comparison. The first column in Fig. \ref{FIG11} is indicative of the deviation in the Lambertian result. As it is visible in the cropped region in Fig. \ref{FIG10} (a), marked in the rectangular part, the beginning and end points of all three folds (marked by A, B and C) should be at about the same level, instead a drastic deviation is showing up at the left side of the surface in results obtained by applying Lambertian reflectance model  as also indicated by the blue area in the corresponding depth map.
\par
However, this deviation is rectified by applying the complete Blinn-Phong model accompanied by PPS as can be seen in the second column of the Fig. \ref{FIG11} and also entirely corrected using this model with PPN approach represented in the third column. Furthermore, three folds of the surface are reconstructed very well in the Blinn-Phong outcomes. This obviously desirable complete reconstruction of those folds cannot be seen in the Lambertian output.

Finally, as also the color alteration in the second row of Fig. \ref{FIG11} shows, high frequency details are recovered as well in the Blinn-Phong outputs especially in the case of PPN approach.

These reconstruction aspects are again clearly observable in another endoscopy image depth reconstruction in Fig. \ref{FIG12} which are the depth resulting from inputs as in Fig. \ref{FIG10} (b). 
Once more, a deviation from the desirable output shape appears in the Lambertian outcome 
especially in the left corner side. 
This part of the surface, which is marked by (C) in the input and 3-D resulting images, has a cavity toward the up-side in reality, which is reconstructed well by the Blinn-Phong outputs in contrast to the Lambertian result.
The latter apparently provides a reconstruction completely on the opposite side for this region of the original surface. 
Let us pay attention also to the second row in Fig. \ref{FIG12}. A curved line of the upper corrugated region (A) is obtained in the right corner of the Blinn-Phong outputs, whereas this region is just a straight line in the right corner of the Lambertian outcome. The height of corrugated regions are obviously more faithfully reconstructed in the Blinn-Phong results compared to the Lambertian one.

Last but not least, it is worth to mention that the viewing angle of the endoscopy cameras is very tight. Using cropped parts of those images in our experiments makes this experiment a highly challenging task of 3D reconstruction. The success of our approach to reconstruct such a tiny range of the depth values without  any knowledge about photographic conditions reveals the capability of our proposed method in challenging real world applications.

In another test with real-world input images, we compared our method with the approach used in \cite {TK05} by making use of the input images depicted as  Fig. 2 (a), Fig. 2 (b) and Fig. 2 (c) in \cite {TK05}. The surface
is a plastic mannequin head, and the plastic material itself shows specularities. 
It is well-known in computer graphics that plastic is a material that can be readily rendered by using the Blinn-Phong model \cite{Pharr10}.

\begin{figure*}[!h]
\vspace{-5mm}
\begin{tabular}{ c }
\hspace{10mm}
\includegraphics[width=0.7\textwidth]{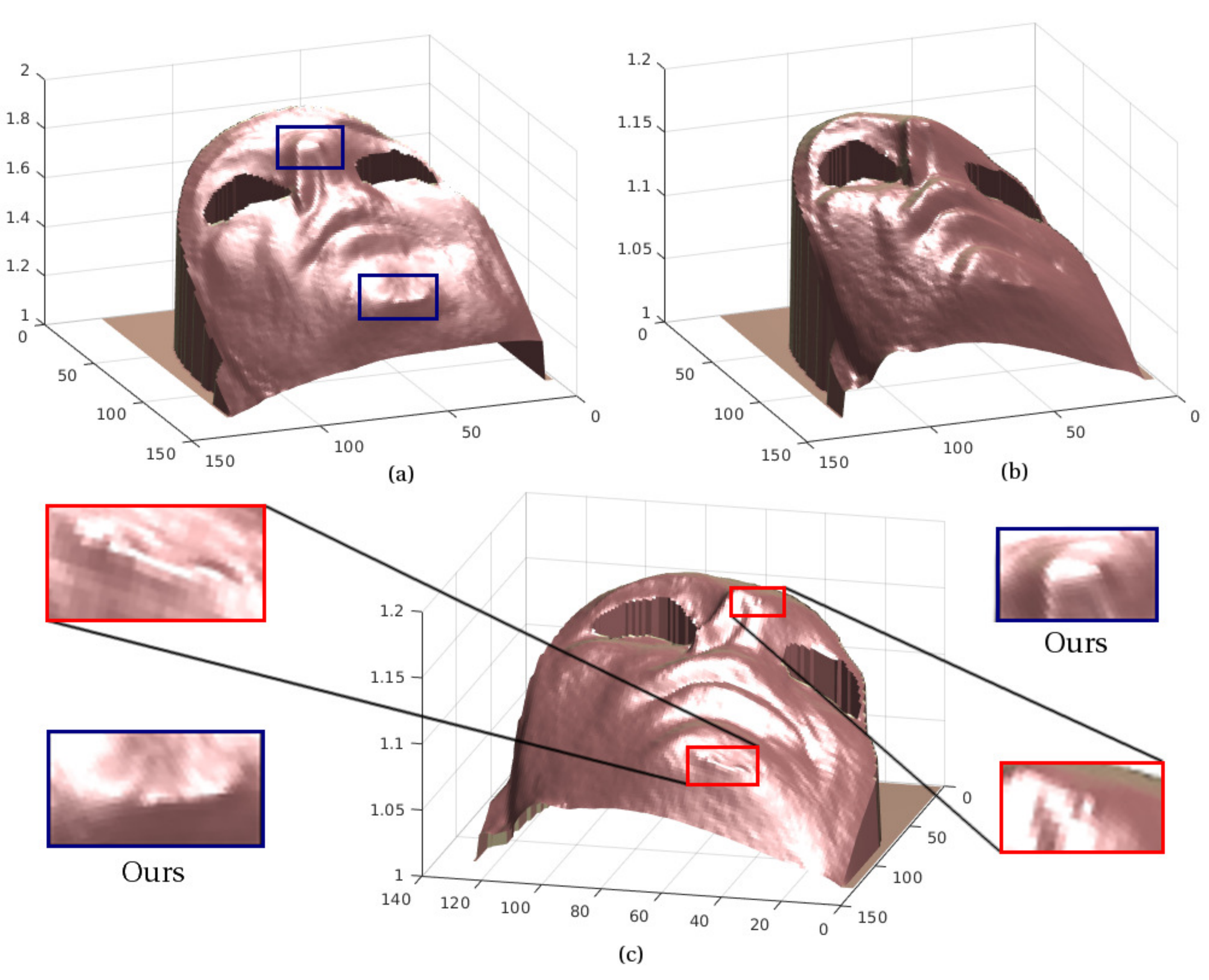}
\end{tabular}
\caption{Depth reconstructions from real world images: 
(a) Results of our proposed method using complete Blinn-Phong model,
(b), (c) results of \cite {TK05}. We have also cropped some parts of our results and shown them together with the same cropped area of outcomes of \cite {TK05} in (c). As it is clear, our approach shows significant superiority over \cite {TK05} in terms of advantages such as smoothness over the rough output of \cite {TK05}, reconstruction success in specularities and absence of deviation.
\label{FIG13}}
\end{figure*}

\begin{figure*}[!h]
\hspace{15mm}
\includegraphics[width=0.85\textwidth]{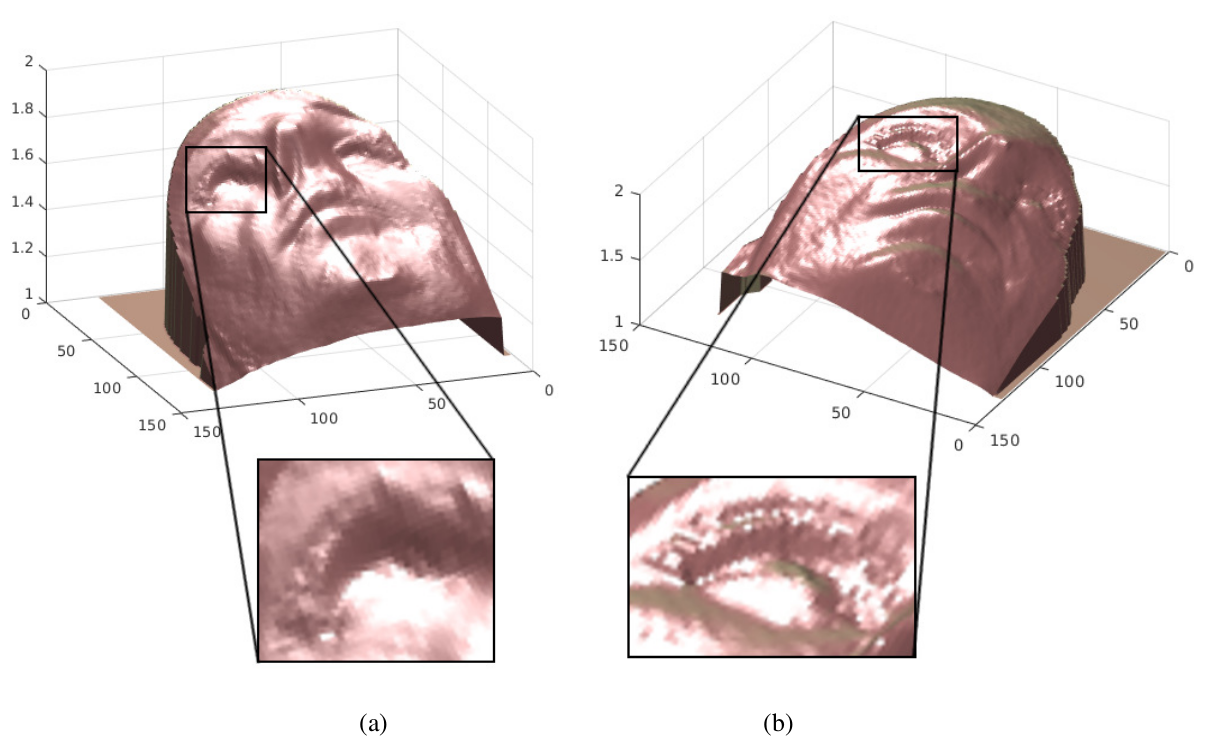}
\caption{Depth reconstructions from real world images: 
(a) results of our proposed method using complete Blinn-Phong model,
(b) results of \cite {TK05}. As it is mentioned in \cite {TK05}, they could not obtain the reconstruction in the presence of eyes (due to the specularities) unlike our approach which provides faithful results even with including eyes.  
\label{FIG14}}
\end{figure*}

The depth reconstructions obtained by our technique and method of Tankus for those real-world images are presented in Fig. \ref{FIG13}. Once again, the deviation from a natural shape in the Lambertian result can be clearly observed in the output in Fig. \ref{FIG13} (b) shown in an identical view with our result in Fig. \ref{FIG13} (a). 
In addition, let us note that the output of the Blinn-Phong model is very clear and smooth, also at highlights. The inhomogeneous recovery of the shape when using the Lambertian model is cropped at some regions such as chin and tip of the nose c.f Fig. \ref{FIG13} (c), where we had to turn the Lambertian result to show these regions. The curved line appearing in the chin and the sharp point at the nose in the Lambertian reconstruction are also visible in \cite {TK05}. Moreover, as proposed in \cite {TK05}, they could not process eyes in images, due to their specularities, while we succeeded in recovering the faithful 3-D shape even with eyes using the complete Blinn-Phong model as presented in Fig. \ref{FIG14}.

\section{Summary and Conclusion}
A new framework in PS considering the complete perspective Blinn-Phong reflectance including strong specular highlights is presented. The advantages of our method over state-of-the-art PS 
methods and also the Lambertian model are proved via a variety of experiments. The model includes a perspective camera projection. Furthermore, two different techniques applied in 
perspective projection are evaluated. In addition, we have also evaluated the modeling of CCD camera. All results are obtained using a minimum necessary number of input images, which is 
an aspect of practical relevance in different applications and makes PS an interesting technique for close to real-time reconstruction, where a minimal set of images is required. We have 
demonstrated experimentally also the merits of our PS model for possible challenging real world applications, where we recover the surface with high degree of details. Let us also 
comment that our computational times are very reasonable i.e. in the order of a few seconds in all experiments. 

Concerning possible limitations, as with all the possible approaches that 
rely on a parametric representation of surface reflectance, the corresponding additional parameters in the reflectance function have to be fixed.
This issue may provide challenging numerical aspects in the optimization. Also, while the Blinn-Phong model gives already reasonable results as we demonstrated, other 
more sophisticated reflectance models may be adequate for handling highly complicated surfaces, which may be a possible issue of future research. 

\CvmAck{This work is supported by the Deutsche Forschungsgemeinschaft 
under grant number BR2245/4--1.
}

\clearpage
\bibliographystyle{CVM}
{\normalsize \bibliography{ref}}

\end{document}